\pdfoutput=1

\documentclass[11pt]{article}

\usepackage[final]{ACL2023}

\usepackage{times}
\usepackage{latexsym}
\usepackage{amsmath}

\DeclareMathOperator*{\argmin}{arg\,min} 

\usepackage[T1]{fontenc}

\usepackage[utf8]{inputenc}

\usepackage{microtype}

\usepackage{inconsolata}

\newcommand{\quotes}[1]{``#1''}
\usepackage{amsmath}
\usepackage{tikz}
 
\usepackage{adjustbox} 
\usepackage{booktabs} 
\usepackage{multirow}
\usepackage{hhline}
\usepackage{xcolor}
\usepackage{makecell}

\usepackage{amsmath} 
\usepackage{algorithm}
\usepackage{algorithmic}
\usepackage{amsfonts}

\usepackage{hyperref}
\usepackage{arydshln}

\usepackage{amssymb}
\usepackage{pifont}
\usepackage{rotating}

\newcommand\blfootnote[1]{
    \begingroup
    \renewcommand\thefootnote{}\footnote{#1}
    \addtocounter{footnote}{-1}
    \endgroup
}

\usepackage{setspace}

\title{Considering Length Diversity in Retrieval-Augmented Summarization}

\author{Juseon-Do$^{1\dagger}$, Jaesung Hwang$^{1\dagger}$, $^*$Jingun Kwon$^1$, \\ \textbf{Hidetaka Kamigaito}$^2$, \textbf{and Manabu Okumura}$^3$ \\
 $^1$Chungnam National University, $^2$Nara Institute of Science and Technology (NAIST) \\
 $^3$Institute of Science Tokyo \\
 {\tt \{doju00,hjs3545\}@o.cnu.ac.kr} \\
 {\tt jingun.kwon@cnu.ac.kr} \\
 {\tt kamigaito.h@is.naist.jp} \\
 {\tt oku@pi.titech.ac.jp}
 \\}

\begin{document}
\maketitle
\begin{abstract}
This study investigates retrieval-augmented summarization by specifically examining the impact of exemplar summary lengths under length constraints, not covered by previous work.
We propose a Diverse Length-aware Maximal Marginal Relevance (DL-MMR) algorithm to better control summary lengths. This algorithm combines the query relevance with diverse target lengths in retrieval-augmented summarization. Unlike previous methods that necessitate exhaustive exemplar-exemplar relevance comparisons using MMR, DL-MMR considers the exemplar target length as well and avoids comparing exemplars to each other, thereby reducing computational cost and conserving memory during the construction of an exemplar pool. 
Experimental results showed the effectiveness of DL-MMR, which considers length diversity, compared to the original MMR algorithm. DL-MMR additionally showed the effectiveness in memory saving of 781,513 times and computational cost reduction of 500,092 times, while maintaining the same level of informativeness.\blfootnote{$^*$ corresponding author}\blfootnote{$^\dagger$ Equal Contribution}

\end{abstract}

\section{Introduction}
Retrieval-augmented generation (RAG) is a promising approach in natural language processing (NLP) because it allows large language models (LLMs) to improve generation quality by leveraging a broader set of information from external resources via in-context learning (ICL)~\cite{NEURIPS2020_1457c0d6,han2022prototypical,guo-etal-2023-prompt,izacard-grave-2021-leveraging,qiu-etal-2022-evaluating,su2022selective,wang2023selfconsistency,shao-etal-2023-enhancing}. Early efforts to retrieve exemplars have focused on a nearest neighbor (NN) method, that compares only query and exemplar relevance~\cite{shin-etal-2021-constrained,rubin-etal-2022-learning}. To further improve performance, exemplar-exemplar relevance comparisons or employing a two-stage approach for the retrieval have been studied~\cite{ye-etal-2023-complementary,guo-etal-2023-prompt,ye-durrett-2023-explanation,margatina-etal-2023-active}.

However, despite the success of previous studies, the impact of summary lengths in the ICL for retrieval-augmented summarization has not been yet explored for better controlling summary lengths. Because better controlling summary lengths can improve summarization performance~\cite{kwon-etal-2023-abstractive,miculicich2023summarization}, we propose to incorporate length diversity to construct a pool for the retrieval. 
We first conducted preliminary experiments to investigate how the exemplars' target summary lengths affect the summarization. Using advanced models such as ChatGPT (GPT-4-turbo-preview),\footnote{\url{https://chat.openai.com/}} the generated summaries closely matched the retrieved target exemplar lengths, that implies that exemplar length information is crucial in retrieval-augmented summarization. 

Our preliminary experiments led us to focus on diverse target length information in the retrieval from a pool of exemplars (\S\ref{subsec:impact_of_exemplar_length}). In this paper, we propose a Diverse Length-aware Maximal Marginal Relevance (DL-MMR) algorithm for retrieving exemplars by considering not only query relevance but also target length diversity. Unlike the previous MMR method~\cite{MMR}, which computes scores for all pairs of exemplars to obtain relevance-based diverse exemplars, DL-MMR simplifies the process by storing only the target lengths. By skipping the scoring of all exemplar-exemplar pairs, DL-MMR additionally lowers computational cost and saves memory for building the pool of exemplars. 

We conducted experiments on three sentence summarization benchmarks: the Google, BNC, and Broadcast datasets. Then, we performed an in-depth analysis to assess the effectiveness of our DL-MMR algorithm, demonstrating its robustness across the datasets with large target length gaps. Our DL-MMR significantly outperformed the NN method, that shows the effectiveness of considering length diversity. Furthermore, DL-MMR was comparable to the MMR retrieval method, while saving the memory of 781,513 times and the computational cost of 500,092 times without losing informativeness. Human evaluation results also showed that considering length diversity is effective for producing informative and concise summaries in retrieval-augmented summarization.\footnote{Our code 
is available at \url{https://github.com/JuseonDo/DL-MMR}.}

\section{Maximal Marginal Relevance}
\noindent \textbf{MMR.}
The NN-based exemplar retrieval approach considers only the relevance between the exemplars and query~\cite{liu-etal-2022-makes}. Although this approach can retrieve the nearest neighbors of mostly similar exemplars, it may limit diversity. To address this issue, MMR selects exemplars that are relevant to the query while being diverse enough using the following equation~\cite{ye-etal-2023-complementary}: 
\begin{align}
\resizebox{.87\linewidth}{!}{$  \arg\max_{q_j \in {D}/{T}} (1-\lambda) \text{Dist}(q, q_j) - \lambda \max_{q_i \in T} \text{Dist}(q_j, q_i) $}, 
\end{align}
where $\lambda$ is to control the balance between relevance and diversity, and Dist denotes similarity. Assuming a given query $q$ and that we have already selected a set of $T = \{q_i\}$ exemplars, we select the next one using the Equation (1).

\noindent \textbf{Diverse Length-aware MMR.}
Although better controlling summary lengths can improve summarization performance~\cite{kwon-etal-2023-abstractive,miculicich2023summarization}, it has not been fully explored yet in 
retrieval-augmented summarization. 
Our preliminary experiments (in Sec. 3.2) demonstrated that generated summaries generally adhere to the retrieved target exemplar lengths, highlighting the importance of exemplar length information in retrieval-augmented summarization, because 
previous summarization methods have not assumed that the desired length is provided. 


For this purpose, we propose the DL-MMR algorithm, that chooses exemplars from the exemplar pool, based on their similarity to a given query, while ensuring sufficient target length diversity among exemplars. Considering length diversity would 
prevent an LLM from adhering to a specific length.
Algorithm~\ref{algo} describes the process of choosing exemplars from the pool in the inference step by utilizing Equation (2) 
instead: 
\begin{align}
\resizebox{.87\linewidth}{!}{$  \argmin_{q_j \in {D}/{T}} (1 - \lambda) \text{Dist}(q, q_j) - \lambda \min_{q_i \in T} \text{Diff}(q_j, q_i) $},  
\end{align}
where $\lambda$ indicates a weight between relevance and length diversity. Diff represents the length difference. We use min-max scaling to convert values from Diff and Dist.

While MMR necessitates scoring all pairs of exemplars within the pool, resulting in a scoring count of $n(n-1)/2$, where $n$ indicates the number of exemplars in the pool~\cite{ye-etal-2023-complementary}, 
DL-MMR calculates only the scoring count for the target length, which is $n$. Since the semantic similarity is a relative measure, we need to calculate all exemplar pair similarities for MMR. However, since the length information is a fixed value, we can immediately obtain it for DL-MMR. This additionally ensures significant memory and computational cost saving. 
However, please note both DL-MMR and MMR require recursive comparisons for exemplars in the inference step.

\begin{algorithm}[t!]
\caption{Diverse Length-aware MMR}
\begin{algorithmic}[1]
\footnotesize 
\REQUIRE exemplar pool $D = \{q_1 \ldots q_n\}$, given test query $q$, the number of exemplar $k$, length difference $Diff$ and semantic distance $Dist$
\ENSURE selected exemplars $T = \{q_1 \ldots q_k\}$
\STATE $\mathbb{S} := [[Diff(q_i, q_j)]]_{q_i, q_j \in D}$ \COMMENT{pairwise length difference between exemplars in $D$}
\STATE $\mathbb{Q} := [Dist(q, q_i)]_{q_i \in D}$ \COMMENT{distance between query and exemplars in $T$}
\STATE $\mathbb{S}$, $\mathbb{Q} := Scale(\mathbb{S}), Scale(\mathbb{Q})$ \COMMENT{min-max scaling to transform values to be between 0 and 1}
\STATE $T := \{\}$
\WHILE{$\lvert T\rvert < k$}
    \STATE $\hat{q} := \text{Equation(2)}$ \COMMENT{get the next exemplar based on Eq (2)}
    \STATE $T.\text{add}(\hat{q})$
\ENDWHILE
\RETURN $T$
\end{algorithmic}
\label{algo}
\end{algorithm}

\section{Experiments}
\subsection{Experimental Settings}
\noindent \textbf{Datasets.} We used three sentence summarization benchmarks: Google (\textbf{Google}), Broadcast (\textbf{Broad}), and BNC (\textbf{BNC})~\cite{filippova-altun-2013-overcoming,Clarke2008GlobalIF}. 
The Google dataset contains automatically created summaries based on the syntactic dependency trees from news headlines and the article's first sentence. The gold compression ratio for the test dataset is 0.45. The Broadcast and BNC datasets consist of human created summaries. The gold compression ratios for the test datasets are 0.76 and 0.72, respectively. Table~\ref{tab:stats} shows the dataset statistics.
\begin{table}[t]
\begin{adjustbox}{width=\columnwidth,center}
\begin{tabular}{ccccccc}\toprule
\textbf{Dataset} & \textbf{Training}   & \textbf{Valid}   & \textbf{Test} & \textit{Avg Src Len} & \textit{Avg Tgt Len} \\
\midrule
Google & 200,000 & 1,000  & 1,000 & 24.4 ($\pm$9.2) & 9.8 ($\pm$3.1) \\
Broad & - &  -  & 1,370 & 19.8 ($\pm$12.8) & 15.59 ($\pm$9.3) \\
BNC & - &  - & 1,629 & 27.9 ($\pm$15.3) & 19.3 ($\pm$10.7) 
\\\bottomrule
\end{tabular}
\end{adjustbox}
\caption{Statistics of datasets. The values in parentheses indicate the standard deviation of both the source and target lengths, respectively.}
\label{tab:stats}
\end{table}

\noindent \textbf{Evaluation Metrics.}
The summary quality was evaluated using F$_1$ scores of ROUGE-1 (R-1), -2 (R-2), and -L (R-L)~\cite{lin-2004-rouge}, as well as the BERT score (BS)~\cite{bert-score}. 
To assess the summary length 
satisfiability, we calculated \textit{$\Delta CR$}, which is the difference between the model-generated and gold compression ratios~\cite{kamigaito-etal-2018-higher,Kamigaito_Okumura_2020}.
 
\noindent \textbf{Implementation Details.} We used \texttt{Llama2-13b} 
\texttt{-chat-hf} \cite{touvron2023llama}, \texttt{Phi-3-Mini}
\texttt{-128K-Instruct}~\cite{abdin2024phi3}, and \texttt{GPT-4}
\texttt{-turbo-preview} \cite{openai2024gpt4technicalreport} as our backbone. 
We used FAISS~\cite{douze2024faiss} to construct a pool and \texttt{bart-large}~\cite{lewis-etal-2020-bart} for measuring semantic distance.
We used 8 exemplars and $\lambda$ performed best in validation. 


\noindent \textbf{Compared Methods.}
The baseline retrieval methods were as follows: \textbf{Zero-shot} does not select exemplars from the pool; \textbf{Random} selects exemplars randomly from the pool; \textbf{NN} selects exemplars based on the nearest neighbor of the query using semantic similarity~\cite{liu-etal-2022-makes}; \textbf{MMR} additionally incorporates relevance-based exemplar-exemplar diversity~\cite{ye-etal-2023-complementary}; and \textbf{DL-MMR} incorporates length diversity. We considered the length by either 
the compression ratio (DL-MMR$_{cr}$), the length in target word count (DL-MMR$_{tgt}$). Since the length in the source can offer diverse target lengths~\cite{kwon-etal-2023-abstractive}, we also considered the source word count (DL-MMR$_{src}$).
For both DL-MMR$_{tgt}$ and DL-MMR$_{cr}$, we used $\lambda = 0.1$. For DL-MMR$_{src}$, we used $\lambda = 0.5$. For MMR, we used $\lambda = 0.5$ on \texttt{Google}.\footnote{Implementation details and validation performances on other datasets for $\lambda$ are in Appendix A.}

\subsection{Impact of Exemplar Lengths}
\label{subsec:impact_of_exemplar_length}
We first examined how exemplar lengths affect retrieval-augmented summarization. 
We used \texttt{Google} as the dataset and tried to generate summaries by giving exemplars with a specific target compression ratio 
or word count. 
The exemplars with the desired target compression ratio or word count were randomly extracted from the pool. Table~\ref{tab:preliminary} shows the results. LLMs relied on the desired target compression ratio or word count in exemplars. These preliminary experiments led us to consider length diversity for retrieval-augmented summarization because typical summarization does not have specific target length information. Furthermore, both Llama-2-13b and GPT-4 faced difficulties when the exemplar lengths or ratios are large.  

\begin{table}[t!]
\renewcommand{\arraystretch}{1.2}
\begin{adjustbox}{width=0.9\columnwidth,center}

\centering
\small
    \begin{tabular}{ccccccccc}
        \toprule
        \multirow{2}{*}{\textbf{\textit{len}}} & \multicolumn{4}{c}{\textbf{Llama-2-13b-chat-hf}} & \multicolumn{4}{c}{\textbf{GPT-4-turbo-preview}} \\
        \cmidrule(lr){2-5}\cmidrule(lr){6-9}
             & \textbf{R-1} & \textbf{R-2} & \textbf{R-L} & \textbf{\textit{gen}} & \textbf{R-1} & \textbf{R-2} & \textbf{R-L} & \textbf{\textit{gen}}\\

\midrule
\textit{5}  & 68.1 & 53.8 & 67.5  & 6.4 & 70.1 & 54.0 & 69.5  & 6.8\\
\textit{10} & \textbf{76.1} & \textbf{64.3} & \textbf{75.2}  & 9.6 & \textbf{75.5} & \textbf{63.5} & \textbf{74.7} &  10.6 \\
\textit{15} & 73.4 & 62.6 & 72.7 &  12.5 & 71.4 & 60.6 & 70.8 &  14.0\\
\textit{20} & 70.4 & 60.3 & 69.7 &  14.8 & 67.7 & 57.4 & 67.1 &  16.3\\
     \midrule
\textit{30\%} & 74.6 & 62.2 & 73.9  & 37\% & 75.1 & 61.7 & 74.3 &  40\% \\
\textit{50\%} & \textbf{75.8} & \textbf{64.0} & \textbf{74.9} &  44\%  & \textbf{75.2} & \textbf{63.2} & \textbf{74.4} &  48\%\\
\textit{70\%} & 73.1 & 62.0 & 72.3  & 54\% & 71.5 & 60.4 & 70.9 &  60\%  \\
\textit{90\%} & 67.7 & 57.3 & 67.0  & 66\% & 66.3 & 56.1 & 65.7 &  74\% \\

\bottomrule
\end{tabular}
\end{adjustbox}

\caption{Affect of exemplar lengths. \textit{len} and \textit{gen} indicate the desired length or ratio, and the generated length or ratio, respectively.}
\label{tab:preliminary}
\end{table}

\begin{table}[t!]
\renewcommand{\arraystretch}{1.2}
\begin{adjustbox}{width=0.95\columnwidth,center}

\centering
\small
    \begin{tabular}{cccccccccc}
    \toprule
    \multirow{2}{*}{\textbf{Data}} & \multirow{2}{*}{\textbf{Method}} & \multirow{2}{*}{\textbf{R-1}} & \multirow{2}{*}{\textbf{R-2}} & \multirow{2}{*}{\textbf{R-L}} & \multirow{2}{*}{\textbf{BS}} & \multirow{2}{*}{\textbf{$\Delta$ \textit{CR}}} & \multicolumn{3}{c}{\textit{\textbf{Cost}}}\\
    \cmidrule(lr){8-10}

    & & & & & & & \multicolumn{1}{c}{\textit{Mem}} & \multicolumn{1}{c}{\textit{Time$_{c}$}} & \multicolumn{1}{c}{\textit{Time$_{i}$}}\\
    \midrule
    \multirow{8}{*}{\textbf{Google}} & \textit{Zero-Shot} & 66.8 & 54.8 & 65.7 & 0.68 & 23.1 & - & - & - \\
    & \textit{Random} & 75.2 & 63.5 & 74.5 & 0.76 & -3.8 & - & - & 0m02s\\
    & \textit{NN} & 78.7 & \underline{67.9} & 77.9 & \textbf{0.79} & \underline{-3.1} & - & - & 17m58s\\\cmidrule{2-10}
    & \textit{MMR} & 78.9 & 68.7 & 78.2 & \textbf{0.79} & -2.8 & 372G & 11h06m & 2h14m\\[0.5ex]
    \cdashline{2-10}\noalign{\vskip 0.5ex}  
    & \textit{DL-MMR$_{cr}$} & 78.0 & 67.3 & 77.3 & 0.78 & -1.5 & 3M & 0m25s & 17m58s\\
    & \textit{DL-MMR$_{tgt}$} & \textbf{79.1} & \textbf{69.0}$^\dagger$ & \textbf{78.5} & \textbf{0.79} & \textbf{-0.7}$^\dagger$ & 476K & 0m00s & 17m58s\\
    & \textit{DL-MMR$_{src}$} & 78.0 & 68.1 & 77.5 & 0.78 & -1.0 & 588K & 0m00s & 17m58s \\
\midrule
  
    \multirow{6}{*}{\textbf{Broad}} & \textit{NN} & \underline{80.1} & \underline{66.2} & \underline{78.8} & 0.77 & \underline{-4.5} & - & - & 0m04s\\\cmidrule{2-10}
    & \textit{MMR} & 80.1 & 65.4 & 78.2 & 0.76 & 4.6 & 25M & 0m28s & 0m17s\\[0.5ex]
    \cdashline{2-10}\noalign{\vskip 0.5ex}  
    & \textit{DL-MMR$_{cr}$} & 78.7 & 64.5 & 77.3 & 0.76 & -6.5 & 28K & 0m00s & 0m04s\\
    & \textit{DL-MMR$_{tgt}$} & \textbf{81.9}$^\dagger$ & \textbf{68.1}$^\dagger$ & \textbf{80.7}$^\dagger$ & \textbf{0.78} & \textbf{0.4}$^\dagger$ & 8K & 0m00s & 0m04s\\
    & \textit{DL-MMR$_{src}$} & 81.5 & 67.6 & 80.4 & 0.78 & -1.8 & 8K & 0m00s & 0m04s\\
    \midrule
  
    \multirow{6}{*}{\textbf{BNC}} & \textit{NN} & \underline{74.5} & \underline{58.8} & \underline{72.1} & 0.69 & \underline{-6.2} & - & - & 0m03s\\\cmidrule{2-10}
    & \textit{MMR} & 75.8 & 59.7 & 73.0 & 0.70 & -1.5 & 18M & 0m22s & 0m14s\\[0.5ex]
    \cdashline{2-10}\noalign{\vskip 0.5ex}  
    & \textit{DL-MMR$_{cr}$} & 73.5 & 57.9 & 71.0 & 0.68 & -8.9 & 20K & 0m00s & 0m03s\\
    & \textit{DL-MMR$_{tgt}$} & \textbf{76.6}$^\dagger$ & \textbf{61.5}$^\dagger$ & \textbf{74.3}$^\dagger$ & \textbf{0.71} & \textbf{0.1}$^\dagger$ & 4K & 0m00s & 0m03s\\
    & \textit{DL-MMR$_{src}$} & 76.0 & 60.8 & 73.6 & 0.70 & -2.6 & 4K & 0m00s & 0m03s\\
    \bottomrule
\end{tabular}
\end{adjustbox}

\caption{Experimental results using zero-shot, random, NN, MMR, and DL-MMR on Llama2-13b-chat-hf. \textit{Mem} denotes the memory required to create the exemplar pool. \textit{Time$_{c}$} and \textit{Time$_{i}$} denote the time spent in constructing and loading exemplars in the inference step, respectively. $\dagger$ denotes the significant improvement (\textit{p}<0.05) compared with NN. We used paired-bootstrap-resampling with 100,000 random samples~\protect\cite{koehn-2004-statistical}.}
\label{tab:pre-phi.} 
\label{tab:main table}
\end{table}

\begin{table}[t!]
\renewcommand{\arraystretch}{1.2}
\begin{adjustbox}{width=0.85\columnwidth,center}

\centering
\small
\begin{tabular}{cccccc}
    \toprule
\textbf{Method} & \textbf{R-1} & \textbf{R-2} & \textbf{R-L} & \textbf{BS} & \textbf{$\Delta$ \textit{CR}}\\

\midrule
NN  & 76.0 & \underline{65.2} & 75.5 & 0.75 & \underline{-4.7}\\
\midrule
MMR & 75.5 & 64.9 & 75.0 & 0.75 & -4.8\\[0.5ex]\cdashline{1-6}\noalign{\vskip 0.5ex}  
DL-MMR$_{cr}$  & 75.3 & 64.6 & 74.8 & 0.74 & \textbf{-2.6}\\
DL-MMR$_{tgt}$ & \textbf{76.8} & \textbf{66.3}$^\dagger$ & \textbf{76.3} & \textbf{0.76} & -2.7$^\dagger$\\
DL-MMR$_{src}$ & 74.2 & 63.1 & 73.5 & 0.73 & -4.8\\

\bottomrule
\end{tabular}
\end{adjustbox}
\caption{Experimental results with Phi-3-mini-128k-instruct on \texttt{Google}. The notations are the same as those in Table~\ref{tab:main table}.}
\label{tab:othermodels}
\end{table}

\subsection{Retrieval-augmented Summarization}
 Table~\ref{tab:main table} shows the performance of Llama-2-13b-chat-hf on \texttt{Google}, \texttt{Broad}, and \texttt{BNC}. For \texttt{Google}, we used the \texttt{Google} training dataset as a pool. For \texttt{Broad} and \texttt{BNC} without their own training dataset, we used \texttt{BNC} and \texttt{Broad} datasets as a pool, respectively. DL-MMR significantly outperformed NN in R-2 and $\Delta CR$. Considering length diversity for the retrieval improves ROUGE scores, though it does not always match the gold length. 
Utilizing the length in target word count outperformed the compression ratio and the length in source word count, which indicates the target length information is crucial in retrieval-augmented summarization. Furthermore, DL-MMR$_{tgt}$ was comparable to MMR while using 781,513 times less memory and being 500,092 times and 7 times faster than MMR in the construction and inference steps on \texttt{Google}, respectively.
Table~\ref{tab:othermodels} shows the performance of the Phi-3-mini-128k-instruct model. 
DL-MMR$_{tgt}$ significantly outperformed both NN and MMR.


\section{Analysis}
\noindent \textbf{Human Evaluation and Case Study.}
We sampled 100 sentences from \texttt{Google} for human evaluation. We assigned 40 evaluators, all of whom have obtained both a US high school and a US bachelor's degree, to rate the results from 1 to 5 (5 is the best) for conciseness (Conc) and informativeness (Infor).
Table~\ref{tab:humeval} shows the results. Considering diverse lengths is essential for producing concise and informative summaries.
Table~\ref{tab:case} shows the retrieved exemplars using DL-MMR${_{tgt}}$ and MMR. It can retrieve exemplars with diverse target lengths. 

\begin{table}[t]
\begin{adjustbox}{width=0.85\columnwidth,center}
\centering
\small
\begin{tabular}{lcccc}
\toprule
       &   \textbf{NN}  & \textbf{MMR}  &\textbf{DL-MMR$_{tgt}$}  & \textbf{Gold} \\\midrule
\textbf{Conc.} & \underline{3.52} & 3.59 & \textbf{3.60}$^\dagger$ & 3.54  \\
\textbf{Infor.} & 3.54 & 3.51 & 3.57 & \textbf{3.60} 
     \\\bottomrule
\end{tabular}
\end{adjustbox}
\caption{Human evaluation results. The notations are the same as those in Table~\ref{tab:main table}.}
\label{tab:humeval}
\end{table}

\begin{table}[t!]
\begin{adjustbox}{width=0.95\columnwidth,center}
\begin{tabular}{l}\toprule
\textbf{Source:} Child mortality rates are dropping but are still high in some parts of the world.\\
\midrule
\textbf{Retrieved Exemplars.} \\
\textbf{1.} \textbf{SRC w/ DL-MMR$_{tgt}$:} Some of the most vulnerable children are still waiting too long\\\hspace{3mm} for adoption placements.\\
\hspace{3mm} \textbf{TGT w/ DL-MMR$_{tgt}$:} Some of the vulnerable children are still waiting too long for\\\hspace{3mm} placements.\\
\hspace{3mm} \textbf{SRC w/ MMR:} Some of the most vulnerable children are still waiting too long for\\\hspace{3mm} adoption placements.\\
\hspace{3mm} \textbf{TGT w/ MMR:} Some of the vulnerable children are still waiting too long for placements.\\

\textbf{2.} \textbf{SRC w/ DL-MMR$_{tgt}$:} Spanish fresh produce exports fell by four per cent year on year\\\hspace{3mm} during the first quarter of 2009.\\
\hspace{3mm} \textbf{TGT w/ DL-MMR$_{tgt}$:} Spanish exports fell.\\
\hspace{3mm} \textbf{SRC w/ MMR:} Cholera is surging again in parts of the world, a World Health\\\hspace{3mm} Organization expert said Thursday, pointing to epidemics in Nigeria and Cameroon.\\
\hspace{3mm} \textbf{TGT w/ MMR:} Cholera is surging in parts of the world.\\

\textbf{3.} \textbf{SRC w/ DL-MMR$_{tgt}$:} Children have gone missing from hospitals in Haiti raising\\\hspace{3mm} fears of trafficking for adoption abroad.\\
\hspace{3mm} \textbf{TGT w/ DL-MMR$_{tgt}$:} Children have gone missing from hospitals in Haiti.\\
\hspace{3mm} \textbf{SRC w/ MMR:} New estimates show the US has the seventh highest cancer rate in the world.\\
\hspace{3mm} \textbf{TGT w/ MMR:} The US has the seventh highest cancer rate in the world.\\

\textbf{4.} \textbf{SRC w/ DL-MMR$_{tgt}$:} The World Bank has warned that world poverty is much greater\\\hspace{3mm} than previously thought.\\
\hspace{3mm} \textbf{TGT w/ DL-MMR$_{tgt}$:} Poverty is greater than previously thought.\\
\hspace{3mm} \textbf{SRC w/ MMR:} Birth rates have dropped for a third year in a row in the United States.\\
\hspace{3mm} \textbf{TGT w/ MMR:} Birth rates have dropped for a third year in a row.\\

\textbf{5.} \textbf{SRC w/ DL-MMR$_{tgt}$:} The American Academy of Environmental Medicine has\\\hspace{3mm} released its latest position paper on electromagnetic field and radiofrequency health\\\hspace{3mm} effects calling for immediate caution regarding smart meter installations.\\
\hspace{3mm} \textbf{TGT w/ DL-MMR$_{tgt}$:} The American Academy of Environmental Medicine has\\\hspace{3mm} released its paper on field and effects calling for immediate caution regarding smart \\\hspace{3mm} meter installations.\\
\hspace{3mm} \textbf{SRC w/ MMR:} One in three children are now living in poverty and the figures are\\\hspace{3mm} set to rise as budget cuts kick in, ministers were warned.\\
\hspace{3mm} \textbf{TGT w/ MMR:} One in three children are now living in poverty.\\

\textbf{6.} \textbf{SRC w/ DL-MMR$_{tgt}$:} British women are more likely to die in childbirth than those\\\hspace{3mm} in the former communist state of Slovenia, new research has shown.\\
\hspace{3mm} \textbf{TGT w/ DL-MMR$_{tgt}$:} British women are more likely to die in childbirth than those\\\hspace{3mm} in the former communist state.\\
\hspace{3mm} \textbf{SRC w/ MMR:} The rapid rise in child obesity may be levelling off, according to figures.\\
\hspace{3mm} \textbf{TGT w/ MMR:} The rise in child obesity may be levelling off.\\

\textbf{7.} \textbf{SRC w/ DL-MMR$_{tgt}$:} World Vision says as of today six million people are affected\\\hspace{3mm} by new flooding in Pakistan.\\
\hspace{3mm} \textbf{TGT w/ DL-MMR$_{tgt}$:} Six million people are affected by new flooding in Pakistan.\\
\hspace{3mm} \textbf{SRC w/ MMR:} Streetism has become one of the major social problems facing humanity\\\hspace{3mm} all over the world.\\
\hspace{3mm} \textbf{TGT w/ MMR:} Streetism has become one of the major social problems facing humanity.\\

\textbf{8.} \textbf{SRC w/ DL-MMR$_{tgt}$:} Birth rates have dropped for a third year in a row in the United States.\\
\hspace{3mm} \textbf{TGT w/ DL-MMR$_{tgt}$:} Birth rates have dropped for a third year in a row.\\
\hspace{3mm} \textbf{SRC w/ MMR:} A Congolese warlord has been jailed for 14 years by the International\\\hspace{3mm} Criminal Court for using child soldiers.\\
\hspace{3mm} \textbf{TGT w/ MMR:} A warlord has been jailed for using child soldiers.\\
\midrule
\textbf{DL-MMR$_{tgt}$:} Child mortality rates are dropping.\\
\textbf{MMR:} Child mortality rates are still high in some parts of the world.\\
\textbf{Gold:} Child mortality rates are dropping.\\
\bottomrule
\end{tabular}

\end{adjustbox}
\caption{Retrieved exemplars and output of Llama-2-13b-chat-hf from \texttt{Google}.}
\label{tab:case}

\end{table}

\noindent \textbf{Impact of Target Length Gaps.}
Since \texttt{Google} has a rather different compression ratio from  \texttt{Broad} and \texttt{BNC} with similar compression ratios, we performed more experiments on \texttt{Broad} and \texttt{BNC} with the \texttt{Google} training dataset as a pool, to investigate the effect of large target length gaps. 
Table~\ref{tab:b-b} shows the results. While retrieval with the use of DL-MMR$_{cr}$ and DL-MMR$_{tgt}$ is effective for summarization on both \texttt{Broad} and \texttt{BNC}, NN, DL-MMR$_{src}$, and MMR encounter difficulties with length generalization. This indicates the importance of considering length diversity for retrieval-augmented summarization. However, $\Delta CR$ was not sufficiently met even when using DL-MMR$_{tgt}$. 
Table~\ref{tab:Gap} shows the results when we separated the test dataset into two, shorter or longer than the average target length (11~\cite{ghalandari-etal-2022-efficient}) in \texttt{Google}. 
Due to the relatively short summaries in \texttt{Google} used for the pool, even DL-MMR$_{tgt}$ encounters difficulties for relatively longer summaries.
The results indicate that further improvements would be desirable 
for constructing a pool by considering target lengths in retrieval-augmented summarization.



\begin{table}[t!]
\renewcommand{\arraystretch}{1.2}
\begin{adjustbox}{width=0.9\columnwidth,center}

\centering
\small
    \begin{tabular}{ccccccc}
    \toprule
    \textbf{Data} & \textbf{Method} & \textbf{R-1} & \textbf{R-2} & \textbf{R-L} & \textbf{BS} & \textbf{$\Delta$ \textit{CR}} \\\midrule

    \multirow{6}{*}{\textbf{Broad}} & \textit{NN} & \underline{67.0} & \underline{52.9} & \underline{65.6} & 0.67 & \underline{-22.8} \\\cmidrule{2-7}
    & \textit{MMR} & 67.5 & 53.4 & 66.1 & 0.67 & -23.8 \\
    [0.5ex]\cdashline{2-7}\noalign{\vskip 0.5ex}  
    & \textit{DL-MMR$_{cr}$} & 71.9 & 57.6 & 70.2 & \textbf{0.71} & -16.51 \\
    & \textit{DL-MMR$_{tgt}$} & \textbf{73.4}$^\dagger$ & \textbf{60.1}$^\dagger$ & \textbf{72.2}$^\dagger$ & \textbf{0.71} & \textbf{-11.8}$^\dagger$ \\
    & \textit{DL-MMR$_{src}$} & 69.2 & 55.7 & 67.9 & 0.69 & -19.0\\
\midrule
    \multirow{6}{*}{\textbf{BNC}} & \textit{NN} & \underline{61.3} & \underline{47.3} & \underline{59.8} & 0.60 & \underline{-27.1} \\\cmidrule{2-7}
    & \textit{MMR} & 60.5 & 46.3 & 58.9 & 0.59 & -27.2 \\[0.5ex]
    \cdashline{2-7}\noalign{\vskip 0.5ex} 
    & \textit{DL-MMR$_{cr}$} & 65.5 & 51.3 & 63.7 & 0.63 &  -19.7\\
    & \textit{DL-MMR$_{tgt}$} & \textbf{67.5}$^\dagger$ & \textbf{53.6}$^\dagger$ & \textbf{65.9}$^\dagger$ & \textbf{0.64} & \textbf{-17.0}$^\dagger$\\
    & \textit{DL-MMR$_{src}$} & 61.7 & 48.0 & 60.2 & 0.60 & -23.5 \\    
    \bottomrule
\end{tabular}
\end{adjustbox}

\caption{Experimental results with Llama-2-13b-chat on \texttt{Broad} and \texttt{BNC} using the \texttt{Google} as a pool.} 
\label{tab:b-b}
\end{table}



\begin{table}[t!]
\renewcommand{\arraystretch}{1.2}
\begin{adjustbox}{width=0.9\columnwidth,center}

\centering
\small
    \begin{tabular}{cccccccc}
        \toprule
            \textbf{Data} & \textit{tgt len} & \textbf{R-1} & \textbf{R-2} & \textbf{R-L} & \textbf{BS} & \textbf{$\Delta$ \textit{CR}} & \textit{cnt}\\

\midrule
\multirow{2}{*}{\textbf{Broad}}
& 0$\sim$11 & \textbf{79.4} & \textbf{64.2} & \textbf{78.4} & \textbf{0.77} & \textbf{-0.6} & 718\\
& 12$\sim$  & 66.8 & 55.5 & 65.4 & 0.66 & -24.1 & 652\\
\midrule
\multirow{2}{*}{\textbf{BNC}}
& 0$\sim$11 & \textbf{76.6} & \textbf{59.6} & \textbf{75.3} & \textbf{0.75} & \textbf{-0.5} & 487\\
& 12$\sim$ & 63.6 & 51.1 & 61.9 & 0.61 & -23.9 & 1,142\\

\bottomrule
\end{tabular}
\end{adjustbox}

\caption{Experimental results with Llama-2-13b-chat using DL-MMR$_{tgt}$. The \textit{cnt} indicates the number of instances within each range.}
\label{tab:Gap}
\end{table}

\noindent \textbf{Impact of Number of Exemplar.}
We conducted further experiments to better understand the impact of  the number of exemplars on performance. 
Table~\ref{tab:numexem} shows the results. We observed that at least four exemplars are required to improve performance while ensuring diversity in retrieval-augmented summarization.\footnote{Additional experimental results are in Appendix \ref{appendix:comparison_to_gpt4}.}

\begin{table}[t!]
\renewcommand{\arraystretch}{1.2}

\centering
\small
    \begin{tabular}{cccccc}
        \toprule
            \textbf{Num} & \textbf{Method} & \textbf{R-1} & \textbf{R-2} & \textbf{R-L} & \textbf{$\Delta$ \textit{CR}}\\

\midrule

    \multirow{5.5}{*}{2} 
    & \textit{NN} & 75.5 & 63.3 & 74.6 & -3.4 \\
    & \textit{MMR}& 75.3 & 62.9 & 74.5 & -2.8 \\
    \cmidrule{2-6}
    & \textit{DL-MMR$_{cr}$}  & 71.6 & 59.6 & 70.8 & 4.0 \\
    & \textit{DL-MMR$_{tgt}$}& 73.2 & 62.2 & 72.4 & 7.4 \\
    & \textit{DL-MMR$_{src}$}& \textbf{75.7} & \textbf{64.2} & \textbf{75.0} & \textbf{3.0} \\
\midrule
    \multirow{5.5}{*}{4} 
    & \textit{NN} & 76.7 & 65.2 & 76.0 & -3.6 \\
    & \textit{MMR}& 77.3 & 65.8 & 76.6 & -3.5 \\
    \cmidrule{2-6}
    & \textit{DL-MMR$_{cr}$} & 76.8 & 65.7 & 76.0 & \textbf{-0.2} \\
    & \textit{DL-MMR$_{tgt}$}& 76.2 & 65.5 & 75.4 & 2.8 \\
    & \textit{DL-MMR$_{src}$}& \textbf{77.2} & \textbf{66.6} & \textbf{76.4} & 0.6 \\

\midrule
    \multirow{5.5}{*}{6} 
    & \textit{NN}& 77.8 & 66.9 & 77.1 & -3.2 \\
    & \textit{MMR}& 77.7 & 66.6 & 77.0 & -3.0 \\
    \cmidrule{2-6}
    & \textit{DL-MMR$_{cr}$} & 78.0 & 67.7 & 77.4 & -1.2 \\
    & \textit{DL-MMR$_{tgt}$}& \textbf{78.3} & 68.0 & \textbf{77.7} &  -0.4 \\
    & \textit{DL-MMR$_{src}$}& 77.8 & \textbf{68.0} & 77.3 &  \textbf{-0.01} \\
\midrule
    \multirow{5.5}{*}{8} 
    & \textit{NN}& 78.7 & 67.9 & 77.9 & -3.1 \\
    & \textit{MMR}& 78.9 & 68.7 & 78.2 & -2.8 \\
    \cmidrule{2-6}
    & \textit{DL-MMR$_{cr}$} & 78.0 & 67.3 & 77.3 & -1.5 \\
    & \textit{DL-MMR$_{tgt}$}& \textbf{79.1} & \textbf{69.0} & \textbf{78.5} & \textbf{-0.7} \\
    & \textit{DL-MMR$_{src}$}& 78.0 & 68.1 & 77.5 & -1.0 \\

\midrule

    \multirow{5.5}{*}{10} 
    & \textit{NN} & 79.0 & 68.9 & 78.5 & -2.9 \\
    & \textit{MMR}& 79.3 & 69.2 & 78.7 & -2.9 \\
    \cmidrule{2-6}
    & \textit{DL-MMR$_{cr}$}  & 78.7 & 68.6 & 78.1 & -2.0 \\
    & \textit{DL-MMR$_{tgt}$}& \textbf{79.3} & \textbf{69.5} & \textbf{78.9} & \textbf{-0.8} \\
    & \textit{DL-MMR$_{src}$}& 78.9 & 69.5 & 78.5 & -1.4 \\



\bottomrule
\end{tabular}

\caption{Experimental results of Llama-2-13b-chat-hf on \texttt{Google} in changing the number of exemplars.}
\label{tab:numexem}
\end{table}

\section{Related Work}

\noindent\textbf{Length Constraint.} Text summarization has gained attention for controlling the output sequence length to produce concise summaries while preserving informativeness, because users often consider desired output lengths~\cite{kikuchi-etal-2016-controlling,takase-okazaki-2019-positional,kwon-etal-2023-abstractive,miculicich2023summarization}. Recently, LLMs have demonstrated remarkable zero-shot task-solving abilities, especially in instruction-based settings~\cite{NEURIPS2020_1457c0d6,Radford2019LanguageMA}. Consequently, numerous studies have leveraged instruction-based approaches to control output sequence length, either by directly specifying the desired length~\cite{juseon-do-etal-2024-instructcmp}, or by incorporating multiple control types such as constraints like greater or smaller than a given value~\cite{jie-etal-2024-prompt}.

\noindent \textbf{Retrieval-Augmented Generation (RAG).} RAG has been recognized as a promising method and has been investigated in various NLP tasks~\cite{lee-etal-2019-latent,izacard-grave-2021-leveraging,rubin-etal-2022-learning,guo-etal-2023-prompt, buettner-kovashka-2024-quantifying}. The core idea is to improve the quality of text generation by conditioning LLMs on carefully selected external exemplars. Preivous studies focused on retrieving the most relevant exemplars, which can cause bias, based solely on query–exemplar similarities~\cite{rubin-etal-2022-learning,liu-etal-2022-makes,shin-etal-2021-constrained}. Alternatively, a recent work considered exemplar-exemplar similarities with MMR~\cite{goldstein-carbonell-1998-summarization} for a better chance to illustrate the required reasoning process~\cite{ye-etal-2023-complementary}.\footnote{Appendix \ref{appendix:other_related_work} introduces other related work.}


\section{Conclusion}
We 
revealed that considering length diversity is crucial for retrieval-augmented summarization. To incorporate target length information, we proposed the DL-MMR algorithm, which allows us to obtain a wider range of exemplars with diverse lengths. Our analysis showed that DL-MMR outperforms MMR, resulting in memory and computational cost saving without losing informativeness.

\section*{Limitations}
While our DL-MMR was designed to better control summary lengths with reducing computational and memory costs, the performance gains might diminish as the number of exemplars decreases for obtaining length diversity. We conducted experiments, and the details are in Appendix D. 

In addition, implementing DL-MMR may entail greater complexity than the NN method. To resolve this issue, we will release our code 
for future studies. Furthermore, while our DL-MMR works effectively in English, it might not be directly applicable to languages not covered by the exemplars in our database, especially those with different syntactic and morphological structures. 
We will extend our DL-MMR to multiple languages in the future to evaluate its robustness. 

\section*{Acknowledgments}
We would like to gratefully acknowledge the anonymous reviewers for their helpful comments and feedbacks. This work was supported by Institute of Information \& communications Technology Planning \& Evaluation (IITP) grant funded by the Korea government(MSIT) (No.RS-2022-00155857, Artificial Intelligence Convergence Innovation Human Resources Development (Chungnam National University)).

\bibliography{anthology,custom}
\bibliographystyle{acl_natbib}

\appendix
\section{Implementation Details and Hyperparameter Selection}
Table~\ref{tab:instruction_format} shows instructions for summarization.

\begin{table*}[t]
\centering
\small
\renewcommand{\arraystretch}{0.6}
\begin{tabular}{cl}
\toprule
\textbf{Task} & \textbf{Instruction}\\
\midrule
\multirow{2}{*}{Sentence Summarization} & Sentence:\textbackslash n\{src\}\textbackslash nSummary of the sentence without the\\
& less important words would be:\textbackslash n \\

\bottomrule
\end{tabular}
\caption{Instruction format. The \quotes{src} indicates the placeholder for a source sentence.}
\label{tab:instruction_format}
\end{table*}

For implementation details, we used NVIDIA RTX A6000. For the decoding step, we did set do$_{samples}$=False, length-penalty=1.0. The CPU used for calculations is an Intel 4th Gen Xeon Scalable Processor (16-core).

Table~\ref{tab:googlevalid} shows the performance of the Llama-2-13b-chat-hf model on the \texttt{Google} validation dataset. Tables~\ref{tab:broad-bnc} and \ref{tab:bnc-broad} show the performance of the Llama-2-13b-chat-hf model on \texttt{Broad} with using the \texttt{BNC} training dataset as the pool and \texttt{BNC} with using the \texttt{Broad} training dataset as the pool, respectively. We selected $\lambda$ based on the best average ROUGE scores for each dataset.

Followings are the computational costs between MMR and our DL-MMR in Table~\ref{tab:main table}.

\noindent \textbf{Memory Usage.}

\begin{itemize}
    \item MMR Memory: 372 GB (or 372,000,000 KB)
    \item DL-MMR Memory: 476 KB
    \item Ratio: 781512.6 
\end{itemize}

\noindent \textbf{Time Spent to Score Similarities.}
\begin{itemize}
    \item MMR: 40007.4 seconds
    \item DL-MMR: 0.08 seconds
    \item Ratio: 500,092.5 
\end{itemize}

\noindent \textbf{Time Spent to Retrieve Exemplars in the Inference Step.}
\begin{itemize}
    \item MMR: 8090.6 seconds
    \item DL-MMR: 1077.7 seconds
    \item Ratio: 7.5 
\end{itemize}

\begin{table}[t!]
\renewcommand{\arraystretch}{1.2}
\begin{adjustbox}{width=0.9\columnwidth,center}

\centering
\small
\begin{tabular}{cccccc}
    \toprule
\textbf{Method} & \textbf{R-1} & \textbf{R-2} & \textbf{R-L} & \textbf{BS} & \textbf{$\Delta$ \textit{CR}}\\

\midrule
Zero-shot & 68.7 & 57.2 & 67.9 & 0.69 & 22.4\\
Random & 76.5 & 64.8 & 75.7 & 0.76 & -4.4\\
NN  & \underline{78.8} & \underline{68.0} & \underline{78.1} & 0.78& \underline{-3.8}\\
\midrule
MMR & 79.4 & 68.8 & 78.7 & 0.78 & -3.9\\
[0.5ex]\cdashline{1-6}\noalign{\vskip 0.5ex} 
DL-MMR$_{cr}$  & 79.2 & 68.6 & 78.7 & 0.78 & \textbf{-2.0}\\
DL-MMR$_{tgt}$ & \textbf{79.9}$^\dagger$ & \textbf{69.5}$^\dagger$ & \textbf{79.2}$^\dagger$ & \textbf{0.79} & -1.6$^\dagger$\\
DL-MMR$_{src}$ & 79.7 & 70.0 & 79.1 & 0.79 & -1.5\\

\bottomrule
\end{tabular}
\end{adjustbox}
\caption{Experimental results of Llama-2-13b-chat-hf on the \texttt{Google} validation dataset with using the \texttt{Google} training dataset as the exemplar pool. The notations are the same as those in Table~\ref{tab:main table}.}
\label{tab:googlevalid}
\end{table}

\begin{table}[t!]
\renewcommand{\arraystretch}{1.2}

\centering
\small
    \begin{tabular}{cccccc}
        \toprule
\textbf{Method} & \textbf{$\lambda$} & \textbf{R-1} & \textbf{R-2} & \textbf{R-L}  & \textbf{$\Delta$ \textit{CR}}\\

\midrule
     NN
     & \textit{0.0} & \underline{80.1} & \underline{66.2} & \underline{78.8} & \underline{-4.5} \\\midrule
     \multirow{10}{*}{MMR}
     
     & \textit{0.1} & 79.8 & 65.2 & 78.2 & -5.6 \\
     & \textit{0.2} & 79.8 & 65.6 & 78.3 & -5.2 \\
     & \textit{0.3} & 79.9 & 65.9 & 78.4 & -5.1 \\
     & \textit{0.4} & 79.4 & 65.4 & 77.9 & -6.0 \\
     & \textit{0.5} & 79.4 & 64.9 & 77.7 & -4.5 \\
     & \textit{0.6} & 79.3 & 64.9 & 77.5 & -4.1 \\
     & \textit{0.7} & 79.4 & 64.9 & 77.6 & -1.5 \\
     & \textit{0.8} & 79.4 & 64.9 & 77.5 & 1.3 \\
     & \textit{0.9} & 80.1 & 65.4 & 78.2 & 4.6 \\
     & \textit{1.0} & 80.1 & 65.4 & 78.2 & 5.5 \\
\midrule
     \multirow{10}{*}{DL-MMR$_{tgt}$}
     & \textit{0.1} & 79.9 & 65.8 & 78.7 & -4.5 \\
     & \textit{0.2} & 80.4 & 66.3 & 78.9 & -3.9 \\
     & \textit{0.3} & 80.7 & 66.6 & 79.5 & -2.8 \\
     & \textit{0.4} & 80.7 & 66.7 & 79.4 & -3.1 \\
     & \textit{0.5} & 80.7 & 66.7 & 79.4 & -1.3 \\
     & \textit{0.6} & 80.6 & 66.5 & 79.3 & -1.8 \\
     & \textit{0.7} & 80.6 & 66.4 & 79.1 & -0.8 \\
     & \textit{0.8} & 81.3 & 67.3 & 80.0 & -1.8 \\
     & \textit{0.9} & \textbf{81.9}$^\dagger$ & \textbf{68.1}$^\dagger$ & \textbf{80.7}$^\dagger$ & \textbf{0.4}$^\dagger$ \\
     & \textit{1.0} & 81.5 & 67.6 & 80.2 & -0.7 \\
\bottomrule
\end{tabular}

\caption{Experimental results of Llama2-13b-chat-hf on \texttt{Broad} with using the \texttt{BNC} training dataset as the exemplar pool. The notations are the same as those in Table~\ref{tab:main table}.}
\label{tab:broad-bnc}
\end{table}

\begin{table}[t!]
\renewcommand{\arraystretch}{1.2}

\centering
\small
    \begin{tabular}{cccccc}
        \toprule
\textbf{Method} & \textbf{$\lambda$} & \textbf{R-1} & \textbf{R-2} & \textbf{R-L}  & \textbf{$\Delta$ \textit{CR}}\\

\midrule
     NN
     & \textit{0.0} & \underline{74.5} & \underline{58.8} & \underline{72.1} & \underline{-6.2}   \\\midrule
     \multirow{10}{*}{MMR}
     
     & \textit{0.1} & 74.6 & 58.7 & 72.1 & -6.1\\
     & \textit{0.2} & 75.0 & 59.1 & 72.6 & -5.7\\
     & \textit{0.3} & 74.6 & 58.9 & 72.1 & -5.8\\
     & \textit{0.4} & 75.1 & 59.4 & 72.6 & -4.8\\
     & \textit{0.5} & 75.4 & 59.7 & 72.9 & -4.3\\
     & \textit{0.6} & 75.1 & 59.5 & 72.6 & -4.2\\
     & \textit{0.7} & 75.4 & 59.3 & 72.7 & -4.2\\
     & \textit{0.8} & 74.8 & 58.6 & 72.0 & -4.6\\
     & \textit{0.9} & 75.4 & 59.4 & 72.6 & -3.3\\
     & \textit{1.0} & 75.8 & 59.7 & 73.0 & -1.5\\
\midrule
     \multirow{10}{*}{DL-MMR$_{tgt}$}
     & \textit{0.1} & 74.7 & 58.9 & 72.3 & -5.1\\
     & \textit{0.2} & 75.5 & 60.0 & 73.2 & -4.3\\
     & \textit{0.3} & 75.5 & 60.1 & 73.2 & -3.6\\
     & \textit{0.4} & 75.7 & 60.6 & 73.5 & -2.6\\
     & \textit{0.5} & 76.0 & 60.9 & 73.7 & -0.9\\
     & \textit{0.6} & \textbf{76.6}$^\dagger$ & \textbf{61.5}$^\dagger$ & \textbf{74.3}$^\dagger$ & \textbf{0.1}$^\dagger$\\
     & \textit{0.7} & 76.3 & 61.4 & 73.8 & -0.2\\
     & \textit{0.8} & 76.2 & 60.8 & 73.4 & -0.2\\
     & \textit{0.9} & 76.3 & 60.8 & 73.6 & -0.4\\
     & \textit{1.0} & 76.3 & 60.0 & 73.4 & 0.3\\
\bottomrule
\end{tabular}

\caption{Experimental results of Llama2-13b-chat-hf on \texttt{BNC} with using the \texttt{Broad} training dataset as the exemplar pool. The notations are the same as those in Table~\ref{tab:main table}.}
\label{tab:bnc-broad}
\end{table}

\section{Comparison to GPT-4}
\label{appendix:comparison_to_gpt4}
Table~\ref{tab:gpt} shows the results on \texttt{Google} with using the \texttt{Google} training dataset as the pool. Our DL-MMR$_{tgt}$ using Llama2-13b-chat-hf, which is relatively small, achieved comparable performance compared to NN$_{gpt4}$, which uses ChatGPT (GPT-4-turbo-preview). This indicates that considering diverse target length information is crucial for producing concise and informative summaries in retrieval-augmented summarization. 

\section{Other Related Work}
\label{appendix:other_related_work}

\noindent\textbf{Sentence Compression.} Sentence compression is the task of generating concise and informative summaries by removing unimportant words while preserving fluency. Following the success of tree trimming \citep{jing-2000-sentence,knight,hori,clarke-lapata-2006-models,berg-kirkpatrick-etal-2011-jointly,filippova-altun-2013-overcoming}, \citet{filippova-etal-2015-sentence,klerke-etal-2016-improving,wang-etal-2017-syntax} demonstrate the effectiveness of end-to-end neural network-based approaches. \citet{kamigaito-etal-2018-higher} introduce recursive attention modules that consider syntactic heads \cite{kamigaito-etal-2017-supervised}, which can be extended to document-level summarization \cite{ishigaki-etal-2019-discourse}, similar to graph neural networks in \citet{kwon-etal-2021-considering} leverage parsed discourse trees \citep{Kobayashi_Hirao_Kamigaito_Okumura_Nagata_2020,kobayashi-etal-2021-improving}. \citet{Kamigaito_Okumura_2020} demonstrate the effectiveness of syntactic recursive attention modules combined with the pre-trained language model BERT \cite{devlin-etal-2019-bert}. Reflecting the success of large language models (LLMs), \citet{juseon-do-etal-2024-instructcmp} highlight the usefulness of LLMs and their ability to control output length in sentence compression.

\begin{table*}[ht!]
\renewcommand{\arraystretch}{1.2}

\centering
\small
    \begin{tabular}{ccccccccccc}
        \toprule
            \multirow{2}{*}{\textbf{Method}} & \multicolumn{2}{c}{\textbf{R-1}} & \multicolumn{2}{c}{\textbf{R-2}} & \multicolumn{2}{c}{\textbf{R-L}} & \multicolumn{2}{c}{\textbf{BERTScore}} & \multicolumn{2}{c}{\textbf{$\Delta$ \textit{CR}}}\\
\cmidrule(lr){2-3} \cmidrule(lr){4-5} \cmidrule(lr){6-7} \cmidrule(lr){8-9} \cmidrule(lr){10-11}

& \multicolumn{1}{c}{valid} & \multicolumn{1}{c}{test} & \multicolumn{1}{c}{valid} & \multicolumn{1}{c}{test} & \multicolumn{1}{c}{valid} & \multicolumn{1}{c}{test} & \multicolumn{1}{c}{valid} & \multicolumn{1}{c}{test} & \multicolumn{1}{c}{valid} & \multicolumn{1}{c}{test}\\

\midrule
     \textit{Zero-Shot} & 68.7 & 66.8 & 57.2 & 54.8 & 67.9 & 65.7 & 0.69 & 0.68 & 22.4 & 23.1 \\
     \textit{Random} & 76.5 & 75.2 & 64.8 & 63.5 & 75.7 & 74.5 & 0.76 & 0.76 & -4.4 & -3.8 \\
     \textit{NN} & 78.8 & 78.7 & 68.0 & 67.9 & 78.1 & 77.9 & 0.78 & \textbf{0.79} & -3.8 & -3.1 \\
     \textit{NN$_{gpt4}$}& 79.8 & \textbf{79.1} & \underline{68.2} & 68.1 & 79.0 & \textbf{78.5} & \textbf{0.79} & \textbf{0.79} & \textbf{-0.7} & \textbf{-0.3}\\
     \midrule
     \textit{MMR}& 79.4 & 78.9 & 68.8 & 68.7 & 78.7 & 78.2 & 0.78 & \textbf{0.79} & -3.9 & -2.8 \\
[0.5ex]\cdashline{1-11}\noalign{\vskip 0.5ex} 
     \textit{DL-MMR$_{cr}$} & 79.2 & 78.0 & 68.6 & 67.3 & 78.7 & 77.3 & 0.78 & 0.78 & -2.0 & -1.5\\
     \textit{DL-MMR$_{tgt}$} & \textbf{79.9} & \textbf{79.1} & 69.5$^\dagger$ & \textbf{69.0} & \textbf{79.2} & \textbf{78.5} & \textbf{0.79} & \textbf{0.79} & -1.6 & -0.7\\
     \textit{DL-MMR$_{src}$} & 79.7 & 78.0 & \textbf{70.0} & 68.1 & 79.1 & 77.5 & \textbf{0.79} & 0.78 & -1.5 & -1.0 \\

\bottomrule
\end{tabular}

\caption{Experimental results based on zero-shot, random, NN, MMR, and DL-MMR based on Llama-2-13b-chat-hf and GPT-4-turbo-preview. $\dagger$ indicates the improvement is significant (\textit{p}<0.05) compared with NN$_{gpt}$.} 
\label{tab:gpt}
\end{table*}

\end{document}